\documentclass[11pt]{article} % For LaTeX2e
\usepackage{rldm,palatino}
\usepackage{graphicx}
\usepackage{wrapfig}
\usepackage{subfigure}

\usepackage{float} 
\usepackage{amsmath,amssymb}
\definecolor{blue1}{rgb}{0.2, 0.2, 0.6}

\usepackage[square,sort]{natbib}

\title{Hierarchical Reinforcement Learning of Locomotion Policies in Response to Approaching Objects: A Preliminary Study}

\author{
Shangqun Yu \\
Department of Computer Science\\
Brown University\\
Providence, RI 02906 \\
\texttt{syu68@cs.brown.edu} \\
\And
Sreehari Rammohan \\
Department of Computer Science\\
Brown University\\
Providence, RI 02906 \\
\texttt{sreehari@brown.edu} \\
\AND
Kaiyu Zheng \\
Department of Computer Science \\
Brown University \\
Providence, RI 02906 \\
\texttt{kzheng10@cs.brown.edu} \\
\And
George Konidaris \\
Department of Computer Science \\
Brown University \\
Providence, RI 02906 \\
\texttt{gdk@cs.brown.edu} \\
}

\begin{document}

\maketitle

\begin{abstract}

  Animals such as rabbits and birds can instantly generate locomotion behavior in reaction to a dynamic, approaching object, such as a person or a rock, despite having possibly never seen the object before and having limited perception of the object's properties. Recently, deep reinforcement learning has enabled complex kinematic systems such as humanoid robots to successfully move from point A to point B. Inspired by the observation of the innate reactive behavior of animals in nature, we hope to extend this progress in robot locomotion to settings where external, dynamic objects are involved whose properties are partially observable to the robot. As a first step toward this goal, we build a simulation environment in MuJoCo where a legged robot must avoid getting hit by a ball moving toward it. We explore whether prior locomotion experiences that animals typically possess benefit the learning of a reactive control policy under a proposed hierarchical reinforcement learning framework. Preliminary results support the claim that the learning becomes more efficient using this hierarchical reinforcement learning method, even when partial observability (radius-based object visibility) is taken into account.
\end{abstract}

\keywords{
locomotion, reactive control, hierarchical reinforcement learning
}

\startmain % to start the main 1-4 pages of the submission.

\section{Introduction}
% Say how amazing animals are
Animals have the ability to command their body---a complex kinematic system---quickly in response to environment stimuli \citep{Conditionedreflexes}.
% Give a slightly more concrete example that motivated us
When being approached by a person from an unexpected direction, a rabbit can immediately run away, even if the rabbit has never seen a person before. When being thrown a rock, a bird can very quickly fly away to avoid being hit, even if the rock is moving very fast and the bird does not notice until near collision.

% Give an overview of Deep RL for locomotion [This is background]
Our work draws a connection to both this observation of nature and the progress of deep reinforcement learning (RL), where we noticed remarkable progress in using deep RL for locomotion \citep{peng2016terrain,peng2020learning,tassa2018deepmind}.
% Give an overview of related work that involve dynamic enviroment objects too
Many of the existing works in deep RL for locomotion control only consider controlling the agent from point to point. Works that do consider controlling an external object (e.g.~dribbling) assume full observability of the object's properties permitted for the application of animation \citep{peng2017deeploco}. Other works present empirical evidence for the viability of training end-to-end locomotion policies directly from pixel input \citep{tassa2018deepmind}, but they typically use third-person images with high sample complexity, which creates challenges for practical implementation. In this work, we focus on learning a reactive locomotion policy, which contrasts the settings in prior works where the agent is typically trained to proactively complete some task.

% Our goal andOverview of the paper
Our goal is to extend the success of deep reinforcement learning for locomotion control to settings where the agent must react to external objects under partial observability.
As a first step, we consider a task where the agent must react to avoid being hit by an approaching ball (the ``Dodge Ball Task''). The only reward signal the agent gets is  a negative reward when being hit by the ball.
%Current challenge
Although reinforcement learning has achieved great improvements in domains such as games \citep{DBLP:journals/nature/MnihKSRVBGRFOPB15} and continuous control for robotics \citep{Gu2017DeepRL}, learning a policy in the continuous control setting with sparse rewards is still a major challenge in RL \citep{DBLP:conf/nips/LiWTZ19}. Hierarchical Reinforcement Learning (HRL), with its structured policy and decomposition of problems into smaller subproblems \citep{levy}, not only has shown strength under these challenging settings, but also provides a solution to reuse low-level skill modules on different tasks \citep{DBLP:conf/nips/LiWTZ19}. Therefore, we investigate the benefits of having prior locomotion experience for this task under an HRL framework.
Our preliminary results have shown that a two level feudal hierarchical agent with pre-trained low-level controller can solve the task with high sample efficiency while the end to end agent completely fails to learn with even five times the training samples.

\section{Hierarchical Reinforcement Learning for Legged Reactive Control}

% \begin{wrapfigure}{r}{0.20\textwidth}
%     \includegraphics[ width=0.2\textwidth]{reactiveControl.png}
%     \caption{In the dodge ball task, a single ball is spawned randomly in the environment (every 300 steps) and shot toward the agent along a linear trajectory. }
%     \label{fig:dodge_ball_task}
% \end{wrapfigure}

A legged reactive control task can be formulated as a partially observable sequential decision-making problem.
The environment state $s$ can be broken into two parts, that is, $s=[s_o, s_h]$, where $s_o$ represents the fully observable internal state of the agent, and $s_h$ represents the state of the external object hidden to the agent. At each time step, the agent executes an action $a$ to control its joint velocities, and it receives an observation, denoted $z= [s_o, \phi(s_h)]$, as a function of the state as a result of the action. The task is specified via a reward function $R(s)$. We are interested in the standard reinforcement learning objective, where the agent needs to maximize the cumulative reward as it interacts with the environment. 

\begin{figure}[h]
    \centering
    \includegraphics[width=1\textwidth]{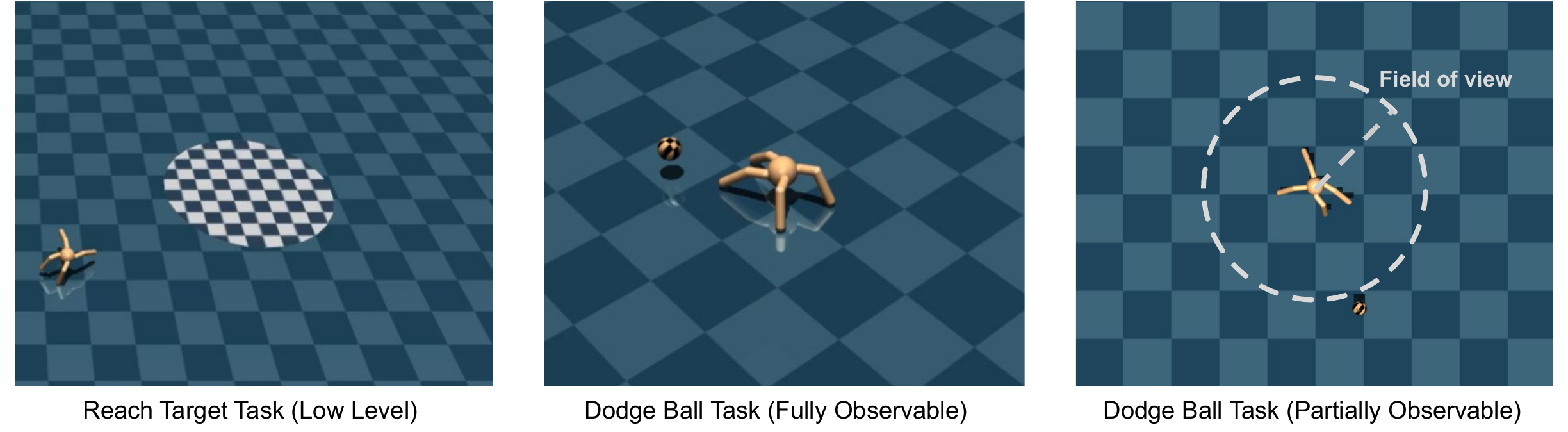}
    \caption{Left: The environment used to train the low-level controller. A target location is defined randomly near the agent (every $100$ steps). The agent will gain reward proportional to the distance it traveled toward the goal. Center: In the dodge ball task, a single ball is spawned randomly in the environment (every $100$ steps) and shot toward the agent along a linear trajectory. The agent receives a $-5$ reward every time it gets hit by the ball. Right: in the partially observable version of the dodge ball task, the agent does not receive information about the ball until it is within the field of view (a circle with $4$ meter radius ).}
    \label{fig:env}
\end{figure}

% In this work we let $n$ denote the number of times the agent collides with the external object. 
In our preliminary investigation of the reactive control setting, the objective is to minimize the expected number of collisions $\mathbb{E}[n]$ by developing an optimal policy $\pi^*(a|z)$ for an agent in a world with a single object, which takes in the current state, and outputs joint velocities for the robot. To this end, we develop the Dodge Ball Task, shown in the center of Figure \ref{fig:env}, implemented in MuJoCo. The agent (an $8$ degree of freedom ant) must learn to dodge a ball which continuously re-spawns from different locations before being shot at the agent along a linear path. In the environment, we define a sparse reward function for our task based on whether a collision happens. Negative reward is only assigned when the robot is hit by the projectile. One episode lasts $1000$ steps and a new ball is spawned randomly in a position that is 5 meters away from the agent and fired at the agent at a speed of 2 meters/second every $100$ steps. An agent following the optimal policy will be able to dodge all balls and thus will have a final cumulative reward close to $0$.

\[R(s) =  \begin{cases}
      $-5$ & \text{agent hit by ball} \\
     $0$ & \text{otherwise} \\
   \end{cases}
\]

\begin{wrapfigure}[25]{r}{0.25\textwidth}
    \includegraphics[width=1\linewidth]{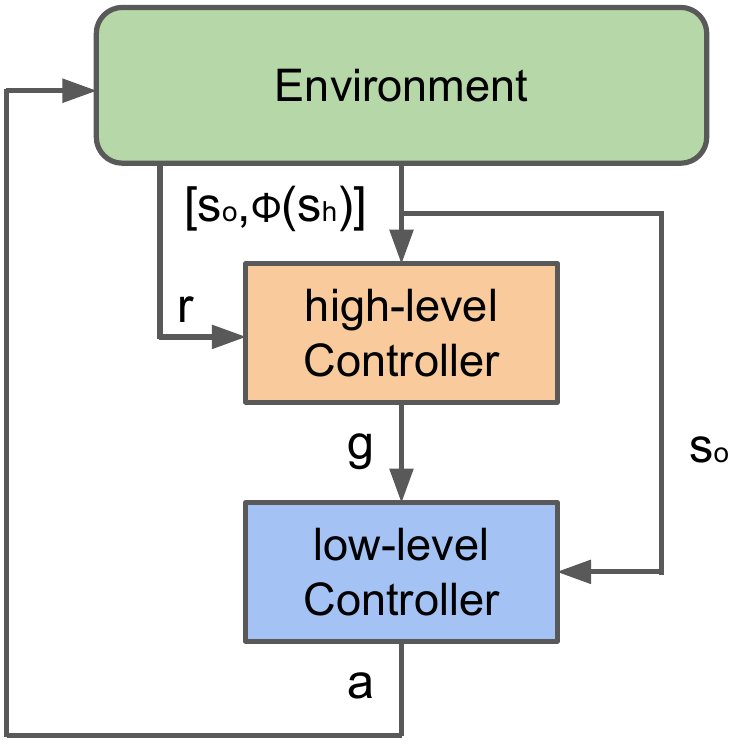}
    \caption{The high-level controller receives a state $s$ from the environment and outputs a subgoal $g$ to the low-level controller consisting of the direction vector it wants the agent to move along. The low-level controller takes in this subgoal and the internal state $s_o$ and outputs a primitive action for execution in the environment.} 
    \label{fig:two_level_hierarchy}
\end{wrapfigure}

We propose a two level Feudal Reinforcement Learning agent shown in Figure \ref{fig:two_level_hierarchy}. Feudal Reinforcement Learning (Feudal RL) is a type of Hierarchical Reinforcement Learning where a high-level controller sets a subtask that is executed by a lower-level controller \citep{NIPS1992_d14220ee,DBLP:journals/csur/PateriaSTQ21}. The state space $S = S_h\times S_o$ consists of the positions and velocities of the agent's joints as well as the position and velocity of the projectile ball. Both high-level and low-level controllers use SAC (soft-actor critic) \citep{haarnoja2018soft} as a base learning algorithm for the task objective.

The high-level controller receives the observation $z = [s_o, \phi(s_h)]$ and outputs a subgoal $g$ to the low-level controller. A subgoal is a 2D point with coordinates relative to the agent. The low-level controller then receives this intent along with the fully observable part of the state $s_o$ and outputs an action consisting of joint velocities for the $8$-DOF agent.

% \begin{wrapfigure}{r}{0.20\textwidth}
%     \includegraphics[ width=0.2\textwidth]{subgoal.png}
%     \caption{The environment used to train the low-level controller}
%     \label{fig:low_level_controller_env}
% \end{wrapfigure}

Similar to how animals possess prior locomotion skills such as running or flying, we first train the low-level controller in a different environment to gain basic locomotion skills for the agent. This environment, shown on the left side of Figure \ref{fig:env}, requires the agent to move along a randomly specified direction vector within $300$ steps, receiving reward proportional to the distance travelled in this direction. The agent receives an observation consisting of the internal configuration (joint positions and velocities) along with a subgoal $g$ with the normalized direction vector. After training, the low-level controller is able to move the agent forward along an arbitrary direction specified by $g$.

Next, the low-level controller is "attached" to the high-level controller in the sense that the high-level controller outputs latent intents (similar to a subgoal direction $g$) which are passed into the low-level controller to compute the primitive action $a$. Note, because the low-level controller is trained agnostic to the high-level task, this module can be re-used -- all that needs to be re-trained is the high-level controller.

% \begin{figure}
%     \centering
%     \includegraphics[width=0.25\textwidth]{arch.pdf}
%     \caption{The high level controller receives a state $s$ from the environment and outputs a latent intent $Z$ to the low-level controller consisting of a direction vector it wants to move along. The low-level controller takes in this latent intent along with the filtered state $\phi(s)$ (filtered state is agnostic to high level task and only contains robot configuration) and outputs a primitive action for the environment.}
%     \label{fig:two_level_hierarchy}
% \end{figure}

\section{Preliminary Experiments}

\subsection{Experiment Setup}
We evaluate the performance of our algorithm in both fully and partially observable settings (Figure~\ref{fig:env}). In the fully observable setting, the agent will receive the complete information of the state $s$, including $[s_o, s_h]$. The full state of the ball $s_h$ is a 6D vector that contains the 3D position (relative to the agent) and velocity of the ball (relative to the world frame). In the partially observable setting, the observation function $\phi_h$ is defined such that the agent can only "see" the ball once it is within a $4$-meter radius (in all other circumstances $0$ is populated for the elements corresponding to the obstacle in the low-dimensional state $s_h$). We use a standard SAC end-to-end trained agent (no low-level controller) as a baseline for comparison. Each agent is trained for $3$ seeds (each lasting $2000$ epochs on both environments). In each epoch, the agent collects $2000$ samples of interaction from the environment and performs $200$ gradient update steps.

\subsection{Results and Discussion}
\begin{figure*}[h]
\centering
\includegraphics[width=0.5\textwidth]{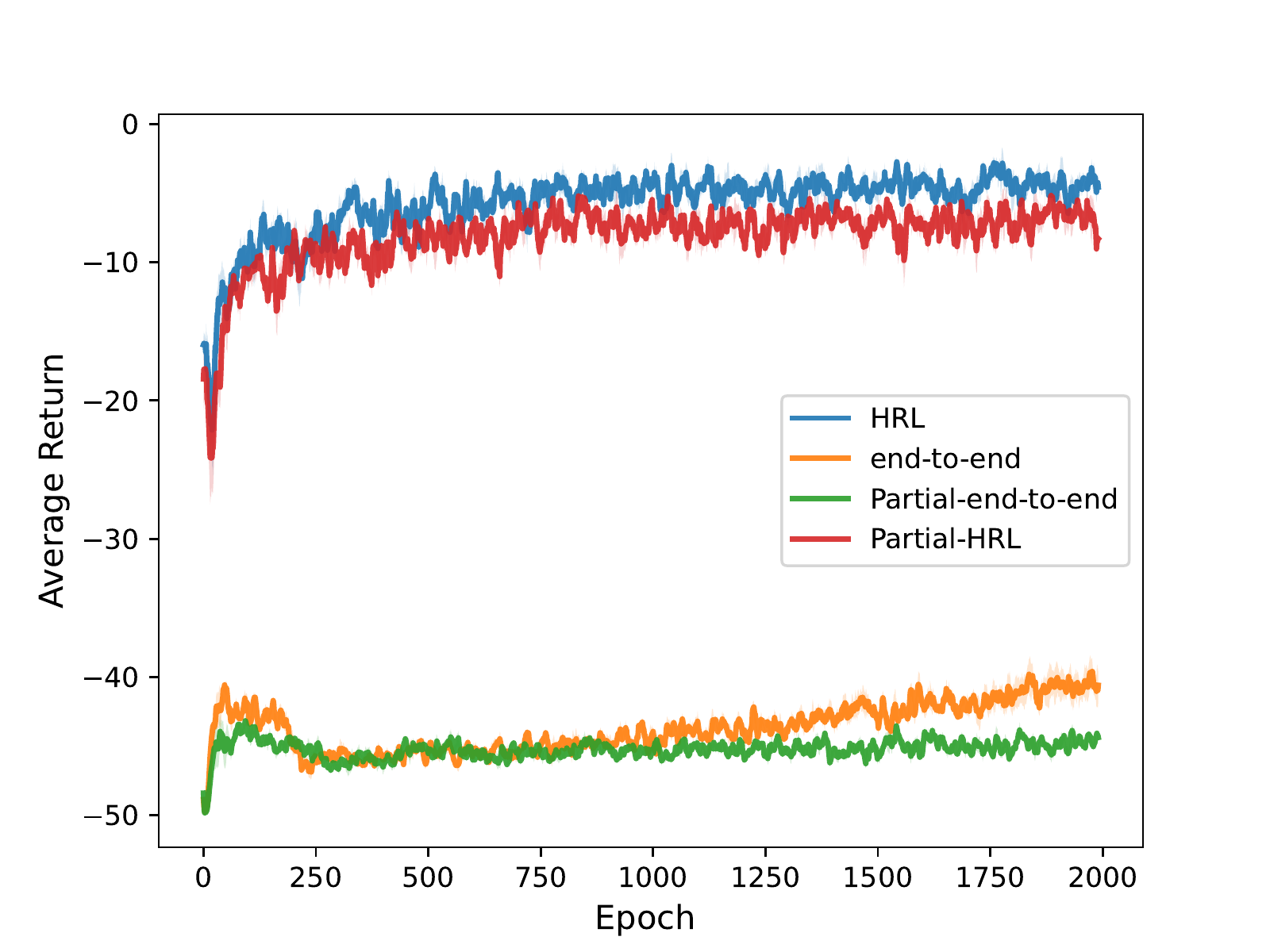}
    \caption{All agents are trained on $3$ seeds. The results show that the HRL agents converged within the first $250$ epochs to a cumulative reward close to $0$, indicating that the agents are able to dodge most balls. Meanwhile, the agents trained end-to-end in both the partially observable and fully observable setting were not able to learn a reactive policy.}
    \label{fig:trainingResults_part_1}
\end{figure*}

\begin{figure*}[h]
\centering

{\small HRL}

\vspace{.1cm}
\includegraphics[width=1\textwidth, trim={0cm 1.cm 0cm 1.5cm}, clip]{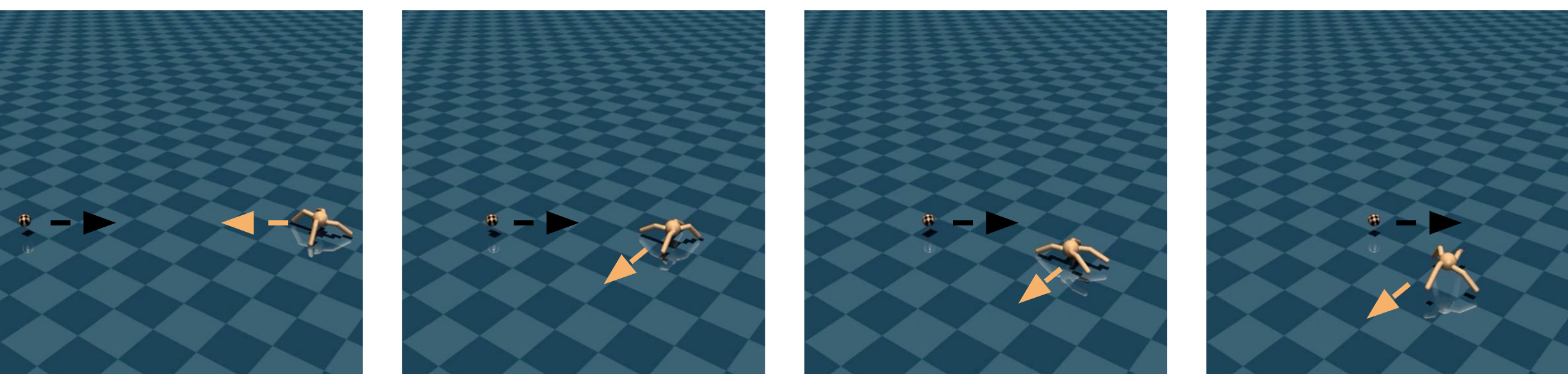}\\

{\small Partial-HRL}%  \vspace{.5cm}

\vspace{.1cm}

\includegraphics[width=1\textwidth, trim={0cm 1.cm 0cm 1.5cm}, clip]{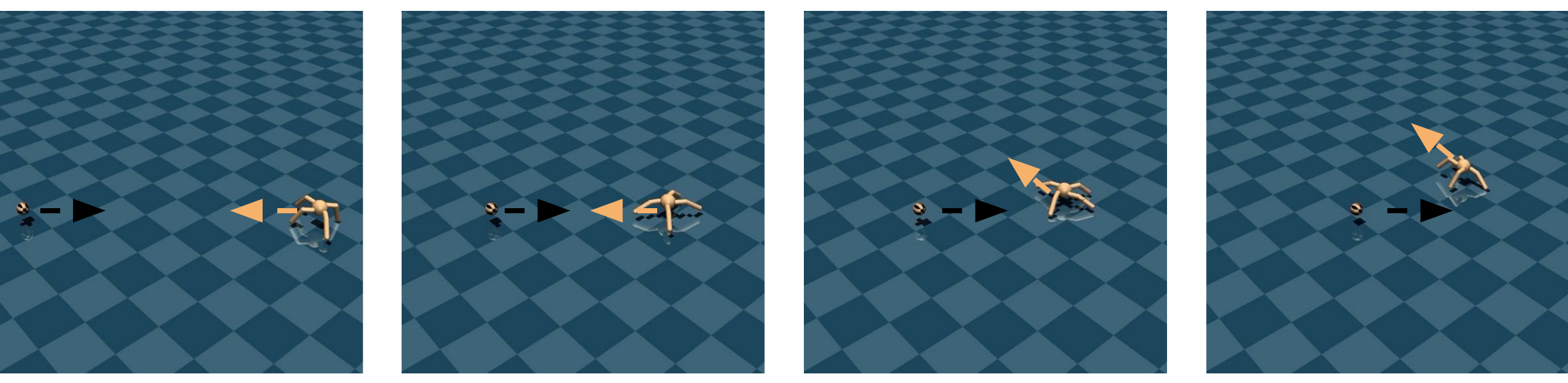}\\

{\small Partial-end-to-end}

\vspace{.1cm}

\includegraphics[width=1\textwidth, trim={0cm 1.cm 0cm 1.5cm}, clip]{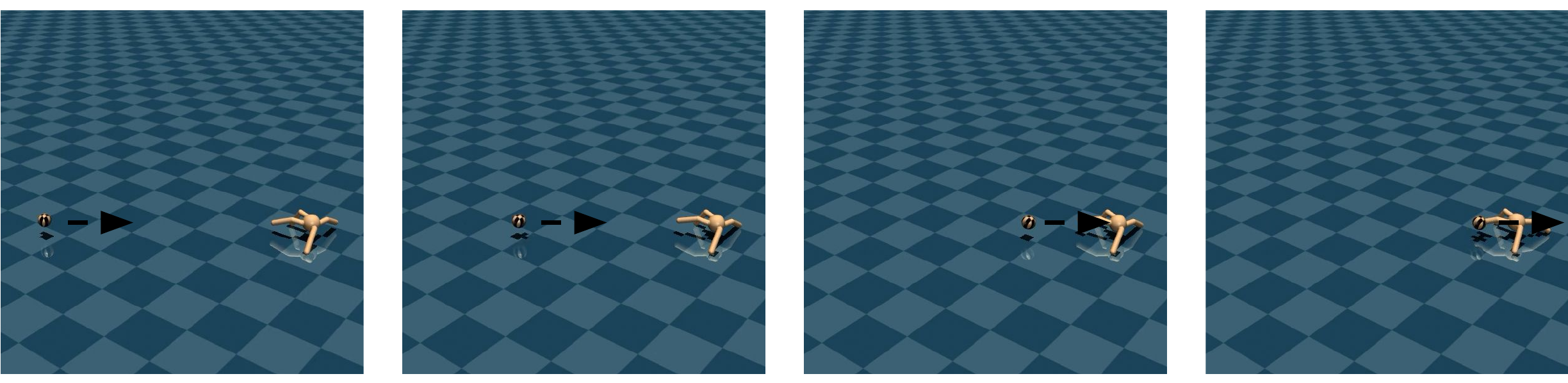}

    \caption{It takes less time for the HRL agent in the fully observable environment (1st row) to react when the ball is approaching compared to the HRL agent in the partially observable environment (2nd row), while the end-to-end agent fails to learn an effective policy in both environments. (3rd row) }
    \label{fig:trainingResults_part_2}
\end{figure*}

Our results in Figure \ref{fig:trainingResults_part_1} show that under sparse reward, feudal HRL agents in both the fully observable and partially observable settings are able to learn reactive behavior well. We observe that the high-level controller learns to assign the direction it wants to go in as a subgoal to the low-level controller, and the low-level controller moves the robot toward the desired direction based on the subgoal, which allows the agent to avoid the ball. It's worth noting that the behavior of the HRL agent is slightly different between the partially observable setting and fully observable setting. Under the fully observable setting, the agent tends to react earlier to the ball when it is approaching compared to the agent under partial observability (Figure \ref{fig:trainingResults_part_2}). This matches our intuition since under partial observability, the agent can not "see" the ball until it is within $4$ meters. As a result, the HRL agent under partial observability converges to a slightly lower average return. End-to-end agents struggle to learn anything meaningful under the sparse reward setting and end up with jittering policies for locomotion after extensive training. 

\subsection{Conclusion and Future Work}
To study the task of reactive control, we developed a ball-dodging environment in MuJoCo that involves a subset of the challenges real world robots face such as learning a locomotion policy and acting under partial observability. We also presented a two level feudal hierarchical reinforcement learning framework for the environment and empirically demonstrated significant improvements over an end-to-end trained RL agent. We plan to extend our HRL framework to tasks that involve a variety of kinematic systems with more realistic assumptions of the agent’s perception capabilities in the partial observable setting (e.g. optical flow). We also hope to increase the complexity of approaching objects (increasing the number and changing the dynamics) to create a more challenging domain. Eventually, we hope to realize our framework on a real quadrupedal robot.

\bibliographystyle{plainnat}
\bibliography{references}
\end{document}